# Deep Active Contours Using Locally Controlled Distance Vector Flow

Parastoo Akbari, Atefeh Ziaei, Hamed Azarnoush

Department of Biomedical Engineering, Amirkabir University of Technology, 424 Hafez Ave., Tehran, Iran

**Abstract**

Active contours Model (ACM) has been extensively used in computer vision and image processing. In recent studies, Convolutional Neural Networks (CNNs) have been combined with active contours replacing the user in the process of contour evolution and image segmentation to eliminate limitations associated with ACM's dependence on parameters of the energy functional and initialization. However, prior works did not aim for automatic initialization which is addressed here. In addition to manual initialization, current methods are highly sensitive to initial location and fail to delineate borders accurately. We propose a fully automatic image segmentation method to address problems of manual initialization, insufficient capture range, and poor convergence to boundaries, in addition to the problem of assignment of energy functional parameters. We train two CNNs, which predict active contour weighting parameters and generate a ground truth mask to extract Distance Transform (DT) and an initialization circle. Distance transform is used to form a vector field pointing from each pixel of the image towards the closest point on the boundary, the size of which is equal to the Euclidean distance map. We evaluate our method on four publicly available datasets including two building instance segmentation datasets, Vaihingen and Bing huts, and two mammography image datasets, INBreast and DDSM-BCRP. Our approach outperforms latest research by 0.59 and 2.39 percent in mean Intersection-over-Union (mIoU), 7.38 and 8.62 percent in Boundary F-score (BoundF) for Vaihingen and Bing huts datasets respectively. Dice similarity coefficient for the INBreast and DDSM-BCRP datasets is 94.23% and 90.89%, respectively indicating our method is comparable to state-of-the-art frameworks.

## 1. Introduction

Active Contour Models (ACM), also called snakes, are deformable parametric curves that have become a very influential tool for image segmentation. Given initial coordinates, ACM converges toward the edges of the region of interest by solving an energy optimization problem based on edge information and shape priors such as boundary continuity and curvature through an iterative process [1].

A major drawback of active contours is their dependence on the internal energy parameters chosen by the user and the chosen external energy functional. The user makes a choice by examining the effect of different contour initializations and energy functional parameters on contour convergence to the object boundary. In other words, active contours initialization and parameters are critical factors in contour convergence. If the contour is initialized far from the true boundaries, it may not converge due to presence of local minima. If the contour is parameterized improperly, inaccurate results could be produced.

Different vector fields, may also lead to different results in contour performance. The first external force introduced for active contours, has a limited capture range and requires the initial contour to be set close to the object. Capture range is the area of the image where external energy can guide the active contour to the boundary. It is also not able to detect concavity [1]. Distance vector field (DVF) based on the Euclidean distance has a higher capture range but fails to detect concavity asdone by traditional snakes [2]. A few vector fields including gradient vector flow (GVF) were then introduced to enhance the capability of the snake to capture concave boundaries [3, 4, 5, 6]. The balloon term has also been proposed as an additional term to snake energy; It pushes the snakes' vertices outwards, by applying a force perpendicular to the contour vertices. The initial position of the contour must be selected inside the target when using the balloon method [7].

Due to the active contours dependence on the mentioned terms, the user interaction is necessary to verify the convergence of the contour towards the object boundary.

In recent studies, deep convolutional neural networks (CNNs) have been combined with active contours replacing the user in the process of contour evolution and image segmentation to eliminate limitations associated with ACM's dependence on parameters of the energy functional and initialization.

In 2018, Marcos et al. presented an end-to-end trainable Deep Structured Active Contours (DSAC), which combines ACMs with CNNs in order to estimate active contour parameters on a per-pixel basis automatically. The major drawback of DSAC is that it does not delineate borders on it's external energy properly. Therefore, DSAC is highly sensitive to initialization. If the initial contour is chosen far from the boundary, active contour evolves to the local minima or collapses to a single point due to gradient invisibility [8].

Several models have then been proposed combining ACMs with CNNs [9, 10]. However, they fail to delineate borders accurately and the problem of manual contour initialization still persists.

In our approach, the internal energy parameters and the balloon term are obtained using the DSAC framework. We propose a new method for extracting external energy and automatic initialization to improve segmentation. To address problems of manual initialization, limited capture range due to local minima in DSAC external energy, and poor convergence to boundary, we propose to train a CNN to predict a ground truth mask from which initialization circle and distance transform is derived. To form a vector field in which every pixel learns a vector to the closest point on the ground truth edge with the magnitude of Euclidean distance map, we use distance transform as a local controlling parameter of distance vector flow.

## 2. Related works

Active contours: The ACMs were first introduced by Kass et al. in 1988 by the name of snakes [1]. The original ACMs had some drawbacks, such as their failure to extract acute concave shapes, sensitivity to initialization, limited capture range, parametrization dependency, topological changes, and the need for user interaction. Thus, much effort has been made in order to increase the robustness of active contours. The balloon force proposed in 1991, causes the curve to inflate like a balloon passing over weak edges and preventing it from collapsing to a single point [7]. Several new external forces have been proposed by subsequent researchers. In 1993, the distance potential force based on the Euclidean distance was introduced to increase the capture range, but it failed to detect concavity [2]. In 1998, Gradient Vector Flow (GVF) was introduced to enhance the capability of the snake based on distance potential. This energy was obtained from the edge map of the image, drawing the contour points towards the edges. Compared to Kass et al.'s model, GVF snake was less sensitive to initialization due to larger capture range and able to detect concavity [3]. In 2006, Boundary Vector Field (BVF) which was obtained from binary boundary map was introduced to increase the capture range and the concavity extraction capability of GVF snake and reduce the computational requirements of generating GVF [4]. In 2008, Magnetostatic Active Contour (MAC), was introduced using a new external force field based on magnetostatics and hypothesized magnetic interactions between the active contour and object boundaries [5]. In 2009, Fluid Vector Flow (FVF) was proposed demonstrating improvements in capture range and concavity extraction capability over previous techniques [6].

Interactive image segmentation combining deep CNNs with ACMs: The first work combining deep CNNs and active contours for image segmentation was proposed in 2016 where a CNN was trained to predict a vector for a patch around a given contour node pointing towards the closest point on the true boundary [11]. Marcos et al. proposed an end-to-end trainable Deep Structured Active Contours (DSAC), which combined ACMs with CNNs in order to estimate active contour parameters on a per-pixel basis using a structured support vector machine (SSVM) hinge loss optimizing for IoU [8]. Cheng et al. proposed Deep Active Ray Network (DARNet), which combined energy maps predicted by a CNN with polar representation of active contours, known as active rays, to prevent contour self-intersection and the energy function computational overhead [9]. Gur et al. proposed a fully differentiable framework, ACDRNet, in which the active contour evolved based on a 2D displacement field predicted by an encoder-decoder architecture. ACDRNet adopted a neural renderer to reconstruct the polygon shape from it's vertices. Loss was computed using the difference between the reconstructed polygon shape and the ground truth mask [10].

In recent years, several works have also combined CNNs with level set segmentation approach. Hoogi et al. proposed a CNN based method for adaptive estimation of level set active contour parameters in which CNN outputs 3 probabilities, inside the lesion and far from it's boundaries, close to the lesion boundaries, or outside the lesion and far from it's boundaries which were then used to calculate the energy parameters [12]. Le et al. proposed a recurrent neural network based level-set framework for semantic segmentation of natural images [13]. Hatamizadeh et al. proposed Deep Convolutional Active Contours (DCAC), a fully automatic framework which adopts CNN to predict level set active contours parameters [14]. Wang et al. proposed Deep Extreme Level Set Evolution (DELSE), an interactive framework in which a backbone CNN and user clicks on the extreme boundary points, were used for level set evolution [15]. Recently Hatamizadeh et al. proposed an end-to-end automatically differentiable and backpropagation trainable framework under the name of Trainable Deep Active Contours (TDACs) [16].

### 3. Our Approach

In this section, we introduce the proposed method. As shown in Figure 1, this study uses two CNNs that are trained simultaneously. Given an input image, one CNN predicts the contour's internal energy parameters and the balloon force using DSAC idea. The other CNN generates a ground truth mask from which, an initialization circle and distance transform are generated. We propose to use distance transform as a local controlling parameter of the distance vector flow to form a vector field pointing from each pixel towards the closest point on the boundary with the magnitude of it's Euclidean distance to the edge. CNN predictions are passed to ACM. Then the active contour evolves towards the target by an iterative energy minimization process.

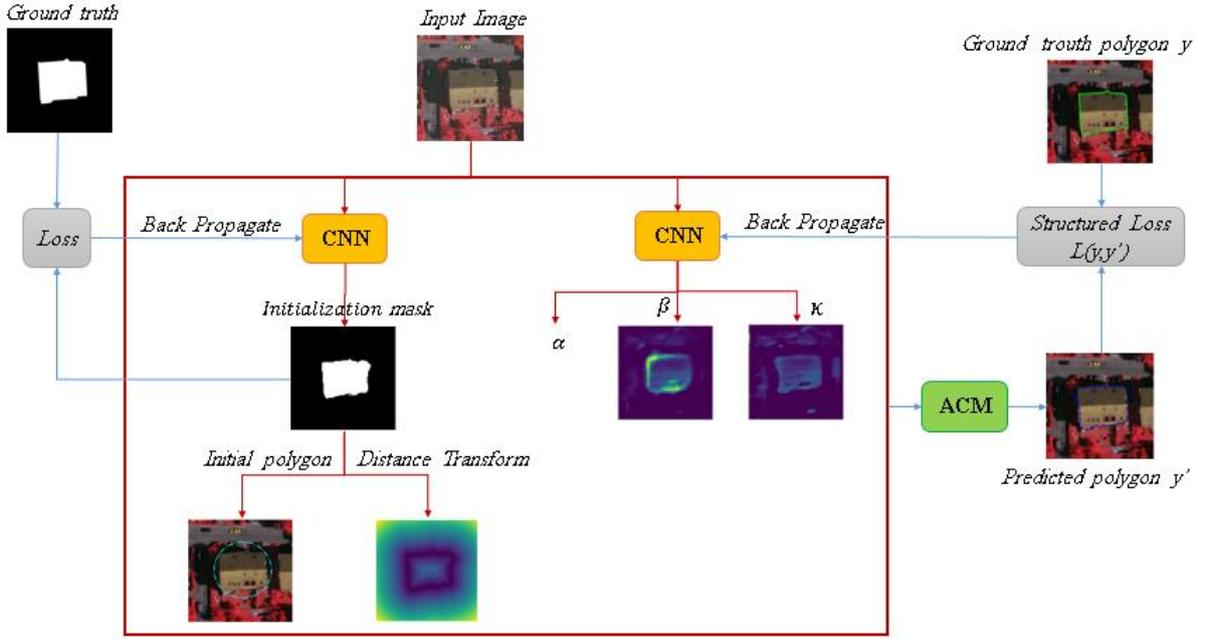

*Figure 1. The proposed framework.*

### 3.1. Contour Presentation

Locally penalized active contours [8] represent classical active contours as a set of polygon points $y_s = (u_s, v_s) \in \mathbb{R}^2$ with s representing the contour node and $s \in \{1, 2, \ldots, L\}$, where L is the number of polygon nodes. The aim of the snake is to evolve towards target edges by minimizing the following energy functional:

$$E(y, x) = \sum_{s=1}^{L}\left[D(x,(y_s)) + \alpha(x,(y_s))\left|\frac{\partial y}{\partial s}\right|^2 + \beta(x,(y_s))\left|\frac{\partial^2 y}{\partial s^2}\right|^2\right] + \sum_{u,v \in \Omega(y)} k(x,(u,v)) \quad (1)$$

where $x \in \mathbb{R}^{U \times V \times d}$ is a notation of input image of size $U \times V$. $D(x), \alpha(x), \beta(x) \in \mathbb{R}^{U \times V}$ are external energy, weight of continuity energy, and weight of curvature energy, respectively. The weighted summation of first and second-order derivatives around the contour is the internal energy penalizing the length and curvature of the polygon, respectively. $\kappa(x)$ is the balloon term and $\Omega(y)$ is the region enclosed by y. We allow β and κ to vary locally, while α is treated as a single scalar for all pixels as proposed in DSAC.

The above formulation makes the contour energy locally adaptive by penalizing each parameter differently at each image pixel. Since no ground truth is available for energy parameters, the problem is defined as a structured support vector machine (SSVM) hinge loss optimizing for IoU. The subgradients of loss $\mathcal{L}$ with respect to the outputs are as follows:

$$\frac{\partial \mathcal{L}(y^i, x^i; \omega)}{\partial \alpha_\omega(x^i)} = \left|\frac{\partial y^i(u,v)}{\partial s}\right|^2 [(u,v) \in y^i] - \left|\frac{\partial \hat{y}^i(u,v)}{\partial s}\right|^2 [(u,v) \in \hat{y}^i] \quad (2)$$

$$\frac{\partial \mathcal{L}(y^i, x^i; \omega)}{\partial \beta_\omega(x^i)} = \left|\frac{\partial^2 y^i(u,v)}{\partial s^2}\right|^2 [(u,v) \in y^i] - \left|\frac{\partial^2 \hat{y}^i(u,v)}{\partial s^2}\right|^2 [(u,v) \in \hat{y}^i] \quad (3)$$

$$\frac{\partial \mathcal{L}(y^i, x^i; \omega)}{\partial \kappa_\omega(x^i)} = [(u,v) \in \Omega(y^i)] - [(u,v) \in \Omega(\hat{y}^i)] \quad (4)$$

where $x^i$, $y^i$ and $\hat{y}^i$ present input image, ground truth contour and predicted contour, respectively and $i \in \{1, 2, \ldots, N\}$. We use the following subgradient to predict the initialization mask:

$$\frac{\partial \mathcal{L}(mapI; \omega)}{\partial mapI_\omega(x^i)} = mapI - Ground\ truth\ mask \quad (5)$$

where $mapI$ is the mask predicted by the CNN.

### 3.2. Locally Controlled Distance Vector Flow (LCDVF)

Contour external energy in DSAC is defined in a way that provides small values on edges of the region of interest and large values elsewhere, the direction of steepest descent $-\nabla D(x) = -\left[\frac{\partial D(x)}{\partial u}, \frac{\partial D(x)}{\partial v}\right]$ moves contour nodes toward the edges [8]. But this external force failed to provide a large capture range for contour, and the initial snake must be defined only inside the region of interest and close enough to the target to evolve. To overcome this issue, we have replaced external energy with distance transform. Distance transform energy can provide a large capture range but cannot guide a snake into boundary concavities. We address this issue by using distance transform itself as a local controlling parameter to multiply by $\nabla D(x)$ preventing contour points from collapsing to a single point resulting in high capture range and eliminating contour sensitivity to initialization. The direction of steepest descent will be as follows:

$$-\nabla D(x) = -DT(x) \times \left[\frac{\partial DT(x)}{\partial u}, \frac{\partial DT(x)}{\partial v}\right] \quad (6)$$

### 3.3. Automatic Initialization

To automatically initialize active contour, the inscribed circle of the initialization mask output by CNN can be found by solving an iterative optimization problem minimizing the difference between the CNN predicted mask and the binary mask of the suggested circle. The circumscribed circle of the mask, can also be found by solving an iterative optimization problem.

Figure 2 and Figure 3 demonstrates parameters predicted by CNN, given an example input image from the Vaihingen data set.

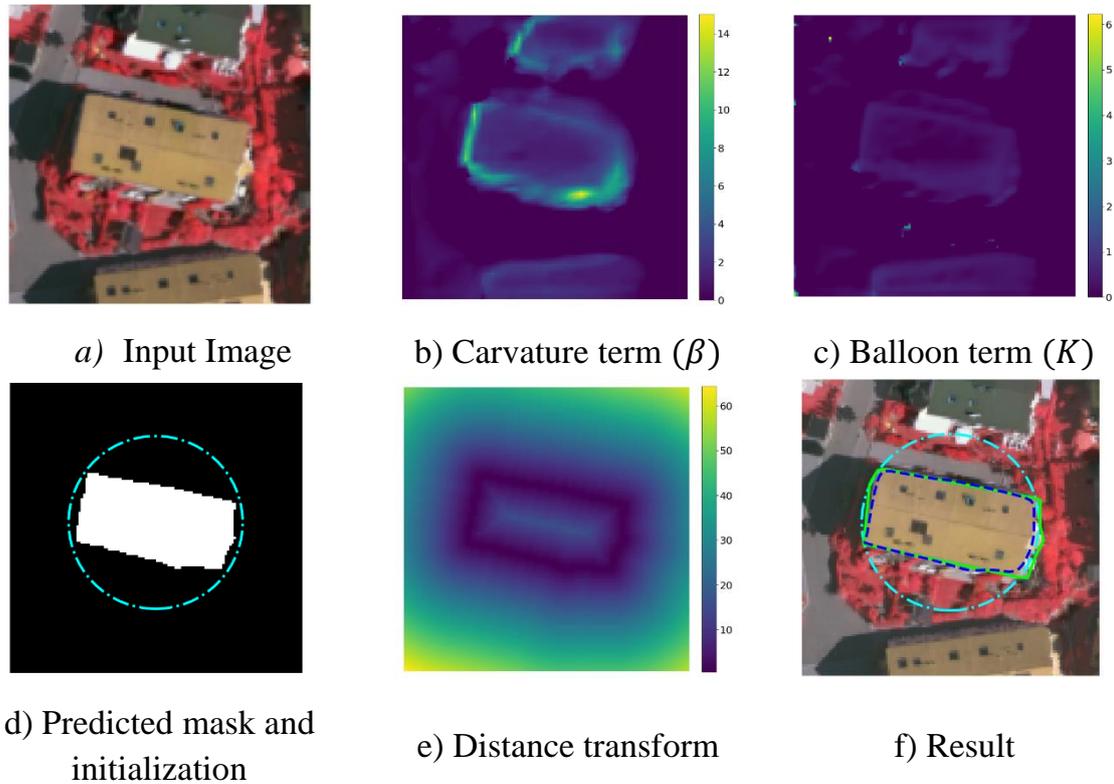

a) Input Image    b) Carvature term ($\beta$)    c) Balloon term ($K$)

d) Predicted mask and initialization    e) Distance transform    f) Result

*Figure 2. Examples of learned parameters for a given input image.*

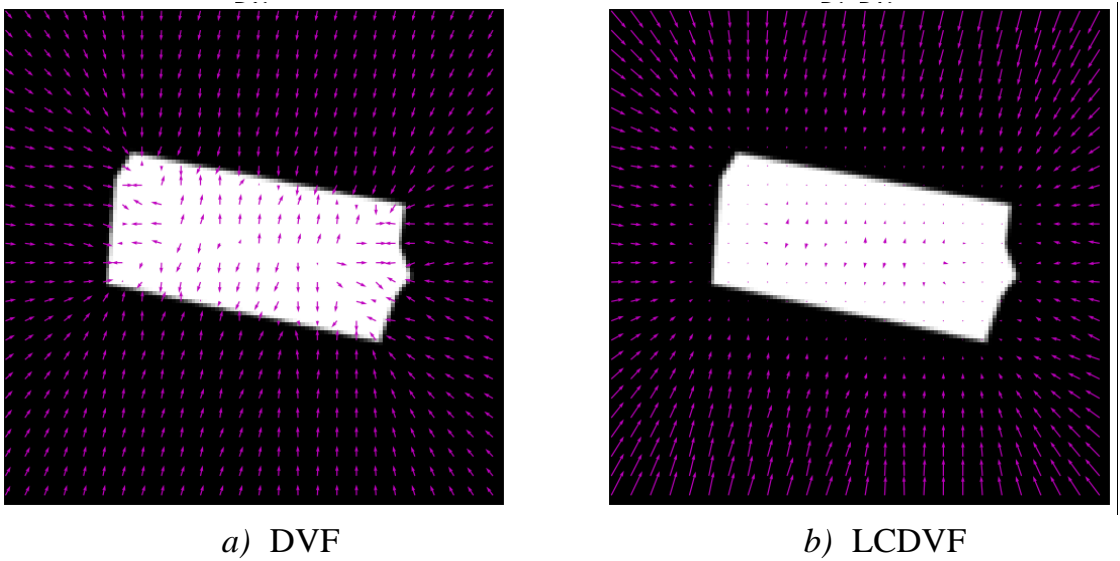

*a)* DVF  *b)* LCDVF

*Figure 3. Examples of learned vector flow.*

### 3.4. CNN Architecture

The backbone CNN structure is illustrated in Figure 4. It is similar to DSAC CNN model [8] and consists of a contracting path with 6 layers with 64, 64, 128,128, 256, and 256 filters, respectively. The size of the first two convolutional layers are 7×7, 5×5, and 3×3 convolution is used for the remaining layers. Each convolutional layer is followed by rectified linear unit (ReLU) activation, batch normalization, and 2×2 max-pooling operation with stride 2 for downsampling. The output tensors of the last four layers are then upsampled to the output size and concatenated. Then, a two-layer MLP with 256 and 64 hidden units is followed by three 1×1 convolution to predict the three output maps, $\alpha(x), \beta(x), \kappa(x)$. The same structure is used to predict an initialization mask to obtain distance transform and estimate the initialization circle. These two CNNs are trained simultaneously.

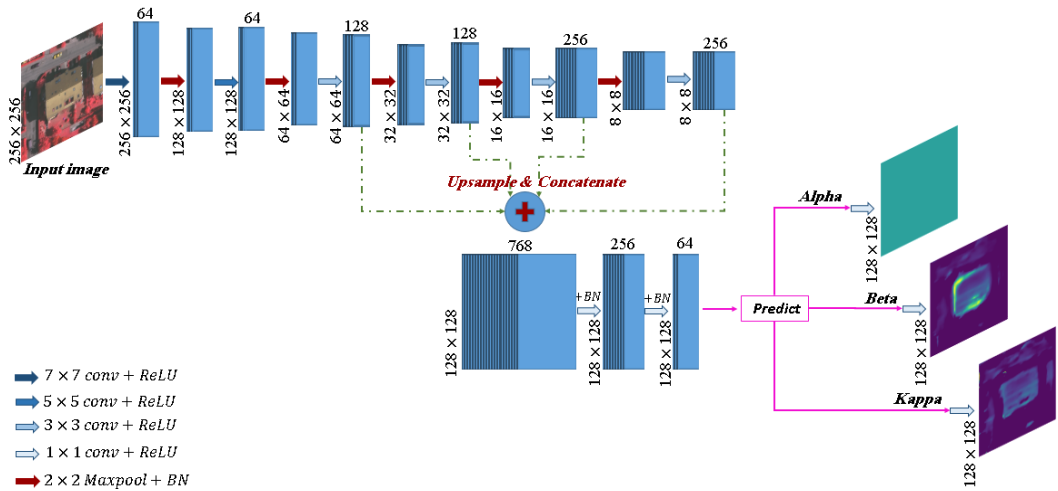

*Figure 4 CNN Architecture.*

## 4. Experiments
### 4.1. Datasets and Augmentation

We evaluate our method on four publicly available datasets: Vaihingen [17], Bing Huts [8], INBreast [18] and DDSM-BCRP (digital database for screening mammography breast cancer research program) [19].

Vaihingen: The Vaihingen buildings dataset consists of 168 building images of size 512×512 pixels at a resolution of 9 cm/pixel in a town in southern Germany. We used the first 100 images for training and the remaining 68 for test as done in previous works [8, 9, 10]. In order to reduce overfitting and improve the generalization capabilities of our model, we have expanded the dataset with data augmentation methods which generate 25 synthetic images from each existing training image by 10 random rotations, 10 random scalings, flipping horizontally and vertically, adding Gaussian Noise, and changing the brightness resulting 2500 training images in total.

Bing huts: The Bing huts dataset consists of 606 challenging images of size 64×64 at a 30 cm/pixel resolution in a rural area of Tanzania. This dataset is divided into 335/271 examples for train/test as done in previous works [8, 9, 10].

INBreast: The INbreast dataset is a mammographic dataset which includes several types of lesions (masses, calcifications, asymmetries, and distortions). The dataset contains a total of 116 accurately annotated mass regions with mass sizes ranging from 15mm$^2$ to 3689mm$^2$, which are divided into 58 ROIs for training and 58 ROIs for test, the same train-test split as previous works [10, 20, 21, 22, 23]. We have increased the number of training images to 870 by 10 random rotation, flipping horizontally and vertically, adding Gaussian Noise, and changing the brightness.

DDSM-BCRP: The DDSM-BCRP dataset is a mammographic dataset selected from the digital database for screening mammography database containing 171 annotated mass regions divided into 84 ROIs for training, and 87 ROIs for test as done in previous works [10, 20, 21, 22].

### 4.2. Evaluation Metrics

To measure the performance of the proposed method, we employ four different metrics, including Intersection over Union (IoU), Dice similarity coefficients (F1 Score) and boundary F-score (BoundF) [9].

IoU metric measures the area of overlap divided by the area of union that is the total number of pixels in both masks excluding the overlap.

$$IoU = \frac{ground\ truth \cap prediction}{ground\ truth \cup prediction} \qquad (7)$$

Dice Coefficient which is closely related to the IoU is twice the area of overlap between the prediction and the ground truth divided by the total number of pixels present across both images.

$$Dice = \frac{2 \times (ground\ truth \cap prediction)}{|ground\ truth| + |prediction|} \tag{8}$$

BoundF is boundary F1score averaged over 1 to 5 pixels thresholds around the boundaries of the ground truth, which demonstrates the model capability in capturing boundaries.

### 4.3. Implementation Details

Our implementation is based on Tensorflow. Input and output sizes are set to 256 and 128 for training Vaihingen and 128 and 128 for other datasets. We trained the network with a batch size of 1 by Adam optimizer. Initial learning rate is set to 0.001 that decays exponentially by a factor of 0.99 every epoch for all datasets except INBreast in which learning rate starts at 0.003 and decays by a factor of 0.96 every 100000 epochs. The number of ACM iterations is set to 50 across all datasets except DDSM-BCRP, for which 10 ACM iterations are found to be sufficient for contour convergence. We use 60 and 100 contour nodes for building and mammography datasets, respectively.

## 5. Results and Discussion
### 5.1. Model capability in reducing overfitting and improving generalization

Figure 5 demonstrates the convergence of Dice coefficient over CNN iterations across all datasets. It can be seen that the proposed method achieves better results as compared to the baseline CNN. Besides, CNNs overfit massively on small datasets. Evaluation metrics over iterations imply combining the capabilities of CNNs with active contour models can both reduce overfitting and improve generalization. Comparing four datasets yields that Vaihingen has the most improvement in generalization and overfitting is almost eliminated.

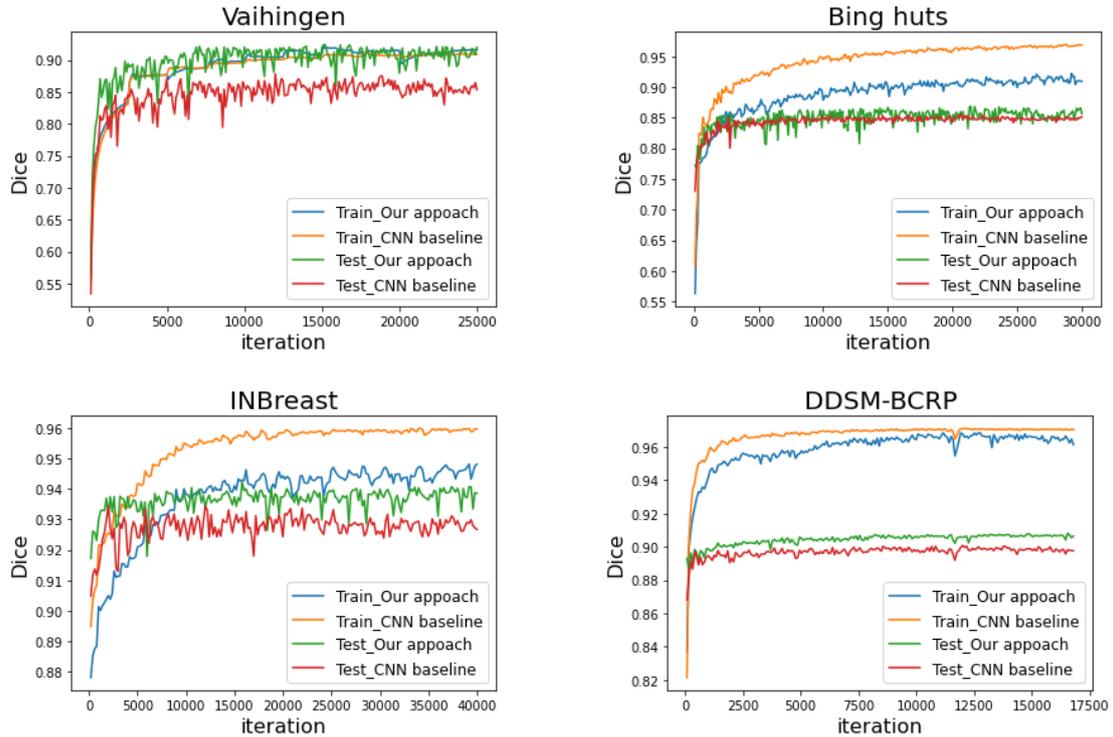

*Figure 5. Dice coefficient over CNN iterations.*

### 5.2. Building datasets segmentation result

We summarize the results of recent methods including DSAC, DARNet, ACDRNet in Table 1 and Table 2 which evaluates our method. These statistics imply our approach outperforms ACDRNet by 0.59 and 2.39 percent in mIoU, 7.38, and 8.62 percent in BoundF for Vaihingen and Bing huts datasets, respectively.

*Table 1. Vaihingen test set results.*

| Method | Vaihingen | | | |
|---|---|---|---|---|
| | mIoU | Dice | BoundF | Initialization |
| FCN | 87.16 | 86.59 | 71.40 | Not applicable |
| DSAC [8] | 71.10 | - | 36.44 | Manual |
| DARNet [9] | 88.24 | 93.65 | 75.91 | Manual |
| ACDRNet [10] | 91.74 | **95.62** | 79.19 | Manual |
| Our approach | **92.33** | 92.44 | **86.57** | **Automatic** |

*Table 2. Bing huts test set results.*

| Method | Bing huts | | | |
| --- | --- | --- | --- | --- |
| | mIoU | Dice | BoundF | Initialization |
| FCN | 85.92 | 85.08 | 66.90 | Not applicable |
| DSAC [8] | 38.74 | - | 37.16 | Manual |
| DARNet [9] | 75.29 | 85.21 | 38.08 | Manual |
| ACDRNet [10] | 84.73 | **91.04** | 58.29 | Manual |
| Our approach | **87.12** | 86.86 | **66.91** | **Automatic** |

Figure 6 shows the proposed vector field, LCDVF, compared to the DSAC energy vector field. In DSAC, the external energy of the contour is updated in a way to provide low values on the boundary and high values elsewhere. It's major drawback is that it does not delineate borders properly; the boundary of adjacent buildings, streets, and cars may be as small as the boundary of the region of interestt, so the contour may converge toward these local minima. In addition, in some parts of the image where the intensity changes are insignificant, DSAC does not provide a good gradient field, making contour points collapse to a single point and failing to properly segment the region. Therefore, DSAC is highly sensitive to the initial location.

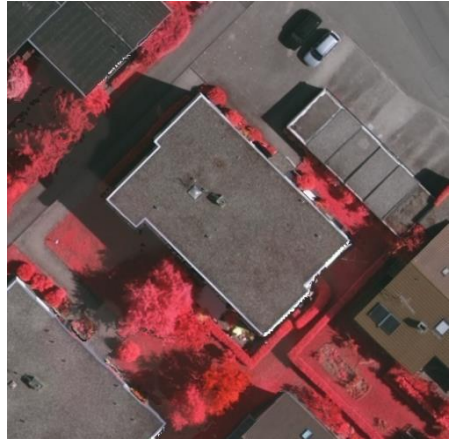
Input image

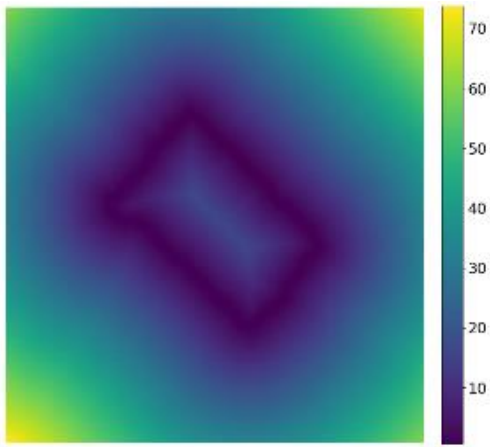
DT

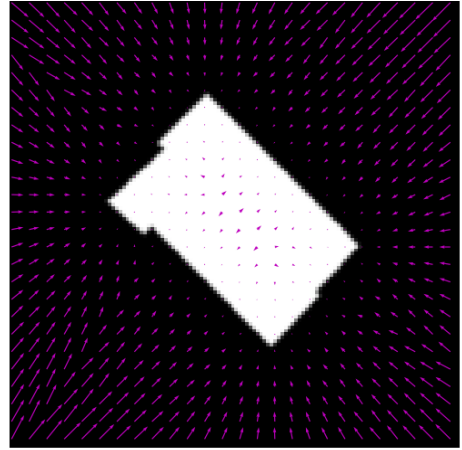
LCDVF

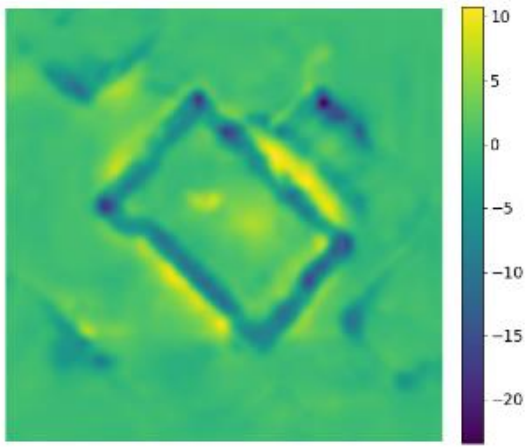
DSAC external energy

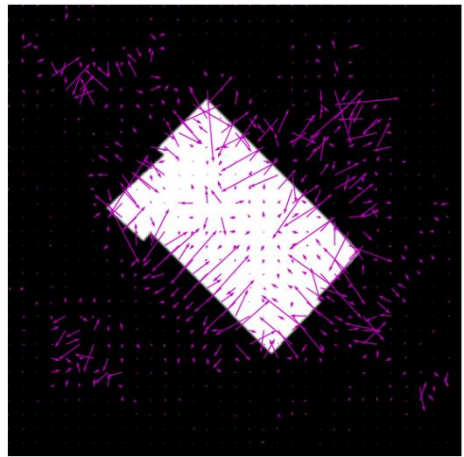
DSAC vector field

*Figure 6. LCDVF compared to DSAC energy vector field.*

The problems of local minima and gradient invisibility in DSAC, are addressed in our proposed approach. Distance vector field with a local controlling parameter equal to the magnitude of the distance transform itself can guide our contour points towards edges with a high capture range. Our method exhibits significant improvement comparing against DSAC, as seen in Figure 7 and Figure 8.

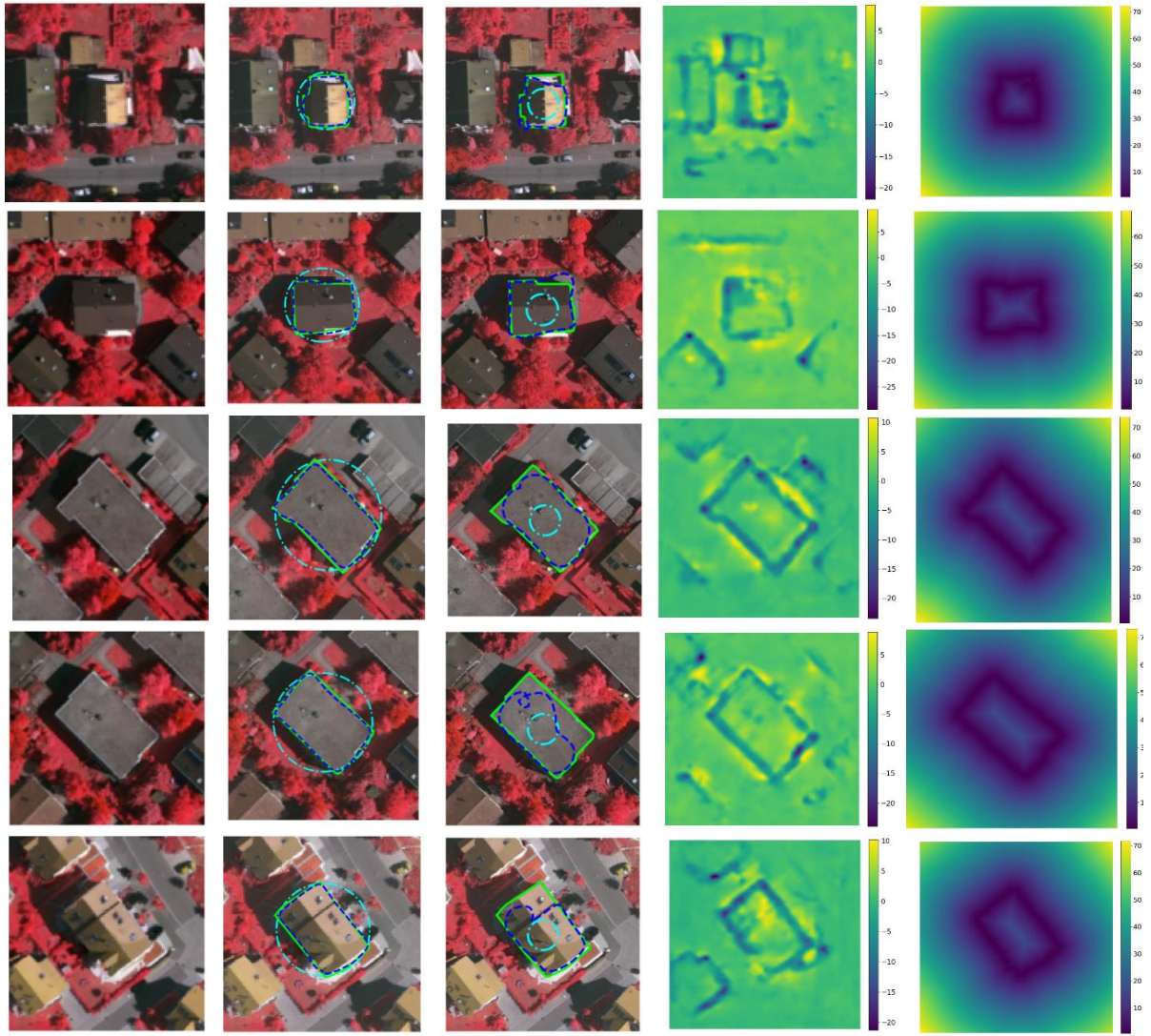

Input    Our approach    DSAC    DSAC Energy    Predicted DT

*Figure 7. Examples of Vaihingen test set. Ground truth, initial contour and the result are in solid green line, dash-dot cyan and dashed blue, respectively.*

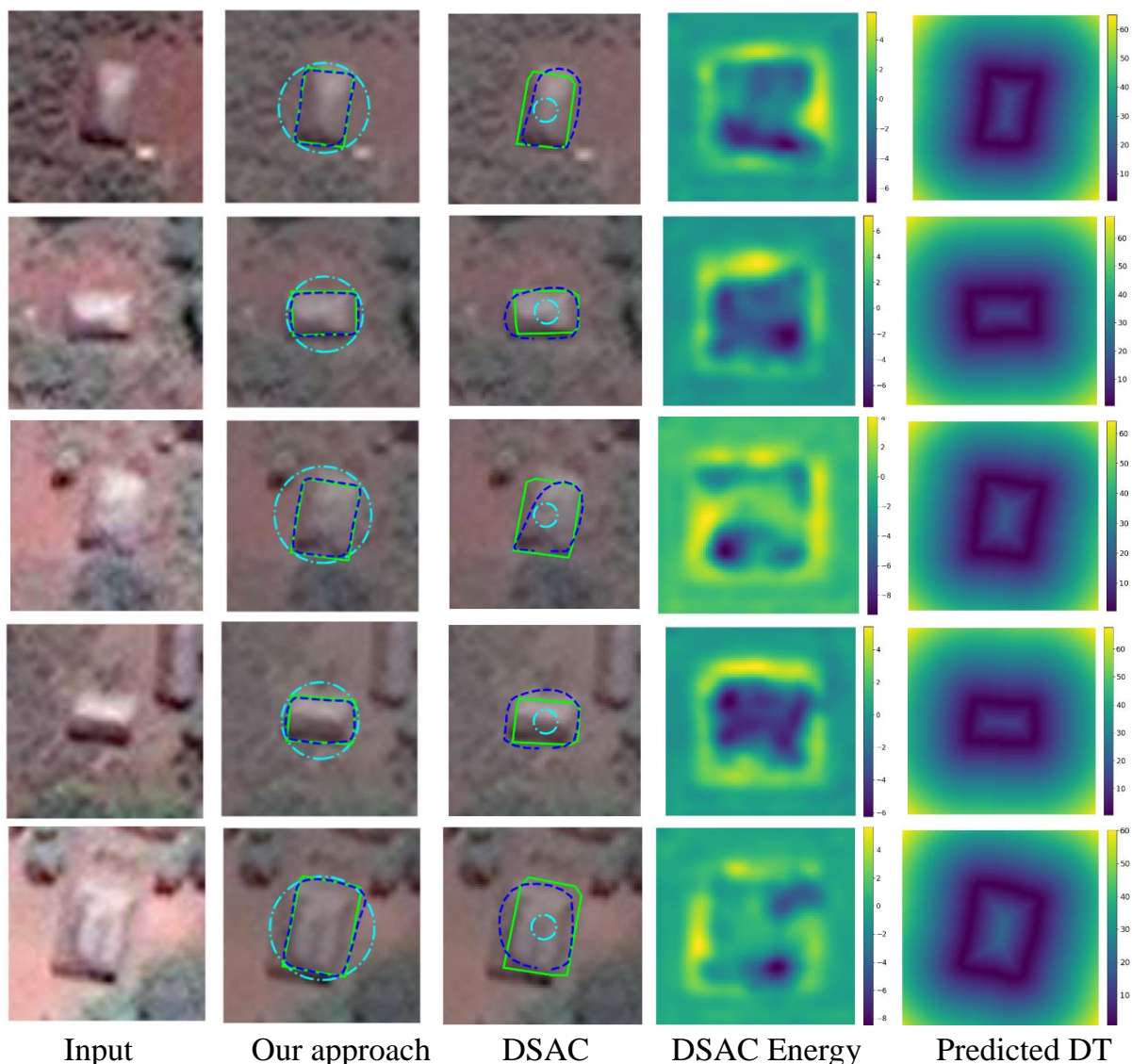

Input     Our approach     DSAC     DSAC Energy     Predicted DT

*Figure 8. Examples of Bing huts test set. Ground truth, initial contour and the result are in solid green line, dash-dot cyan and dashed blue, respectively.*

### 5.3. Mammography datasets segmentation result

Table 3 shows Dice results of the proposed model in comparison with recent methods in mammographic images. Our method Dice is almost equal to the best reported result for INBreast dataset and has a comparable performance compared to the others for DDSM-BCRP. We show sample segmentations in Figure 9.

*Table 3. Medical test set results.*

| Method | INBreast | | DDSM-BCRP | |
|---|---|---|---|---|
| | Dice | Initialization | Dice | Initialization |
| FCN | 93.06 | Not applicable | 89.82 | Not applicable |
| Ball & Bruce [20] | 90.90 | Not applicable | 90.00 | Not applicable |
| Zhu et al. [21] | 90.97 | Not applicable | 91.30 | Not applicable |
| Li et al. [22] | 93.66 | Not applicable | 91.14 | Not applicable |
| Singh et al. [23] | 92.11 | Not applicable | - | Not applicable |
| ACDRNet [10] | **94.28** | Manual | **92.32** | Manual |
| Our approach | 94.23 | **Automatic** | 90.89 | **Automatic** |

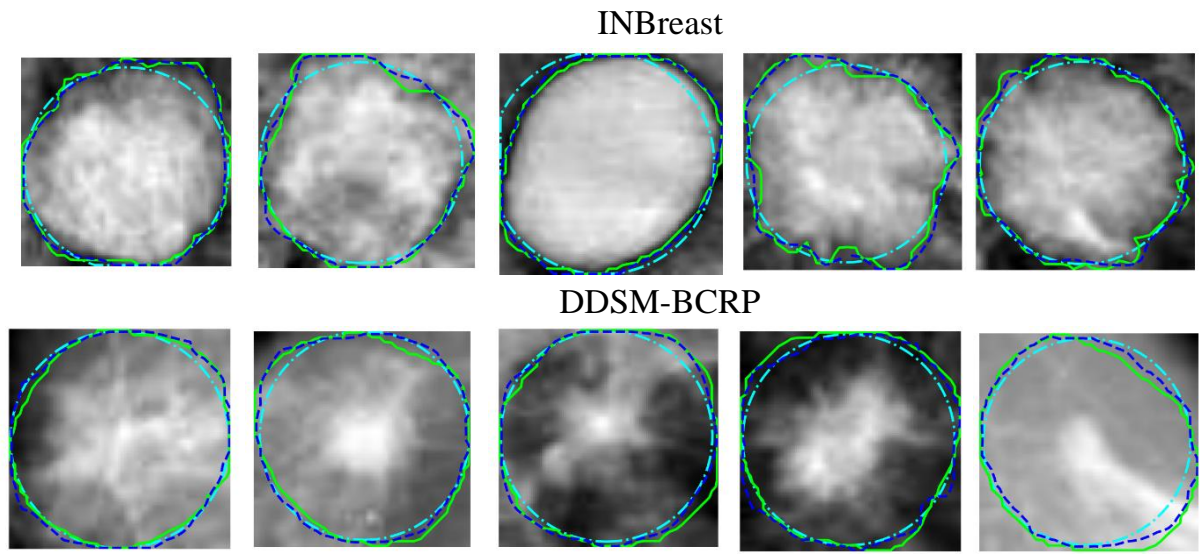

*Figure 9. Examples of Medical data test set. Ground truth, initial contour and the result are in solid green line, dash-dot cyan and dashed blue respectively.*

### 5.4. Model sensitivity
#### 5.4.1. Energy parameters

The importance of energy weighting parameters, the balloon term and the external energy in contour convergence to the boundary is shown in Table 4 and Table 5. Camparing our approach using DVF and LCDVF indicates that using DT as a coefficient to control the size of DVF improves the result significantly across all datasets. BoundF is increased by 3.11%, 0.04%, 7.16%, and 1.44% in Vaihingen, Bing huts, INBreast, and DDSM-BCRP datasets, respectively. Besides, segmentation results obtained by LCDVF, are comparable to state-of-the-art methods indicating that LCDVF alone is very well able to guide the contour to the boundary. As can be seen, LCDVF outperforms our approach in the Bing huts dataset. The balloon term plays an important role in contour convergence. As can be seen, BoundF improved by 6.2%, 0.13%, 5.2% and 1.31% in Vaihingen, Bing huts, INBreast and DDSM-BCRP datasets, respectively, when using balloon energy.

*Table 4. Model sensitivity to energy terms in building datasets.*

| Method | Vaihingen | | | Bing huts | | |
|---|---|---|---|---|---|---|
| | mIoU | Dice | BoundF | mIoU | Dice | BoundF |
| Our approach (No kappa) | 90.23 | 90.20 | 80.37 | 87.07 | 86.79 | 66.78 |
| LCVDF | 90.18 | 90.15 | 79.75 | **87.66** | **87.45** | **68.97** |
| Our approach (DVF) | 91.28 | 91.33 | 83.46 | 87.07 | 86.74 | 66.87 |
| Our approach (LCDVF) | **92.33** | **92.44** | **86.57** | 87.12 | 86.86 | 66.91 |

*Table 5. Model sensitivity to energy terms in mammography datasets.*

| Method | INBreast | | | DDSM-BCRP | | |
|---|---|---|---|---|---|---|
| | mIoU | Dice | BoundF | mIoU | Dice | BoundF |
| Our approach (No kappa) | 80.71 | 93.29 | 55.47 | 72.77 | 90.52 | 41.89 |
| LCVDF | 80.99 | 93.31 | 56.44 | **72.99** | 90.59 | 42.09 |
| Our approach (DVF) | 79.87 | 92.97 | 53.51 | 72.69 | 90.77 | 41.76 |
| Our approach (LCDVF) | **82.72** | **94.23** | **60.67** | 72.16 | **90.89** | **43.20** |

### 5.4.2. Snake Iteration

Figure 10 shows evaluation metrics over the number of snake iterations across all datasets. It is visible that our model performance slightly changes by increasing the number of snake iterations. We choose 50 iterations for Vaihingen, Bing huts, and INBreast and 10 for DDSM-BCRP dataset as the optimal iteration number.

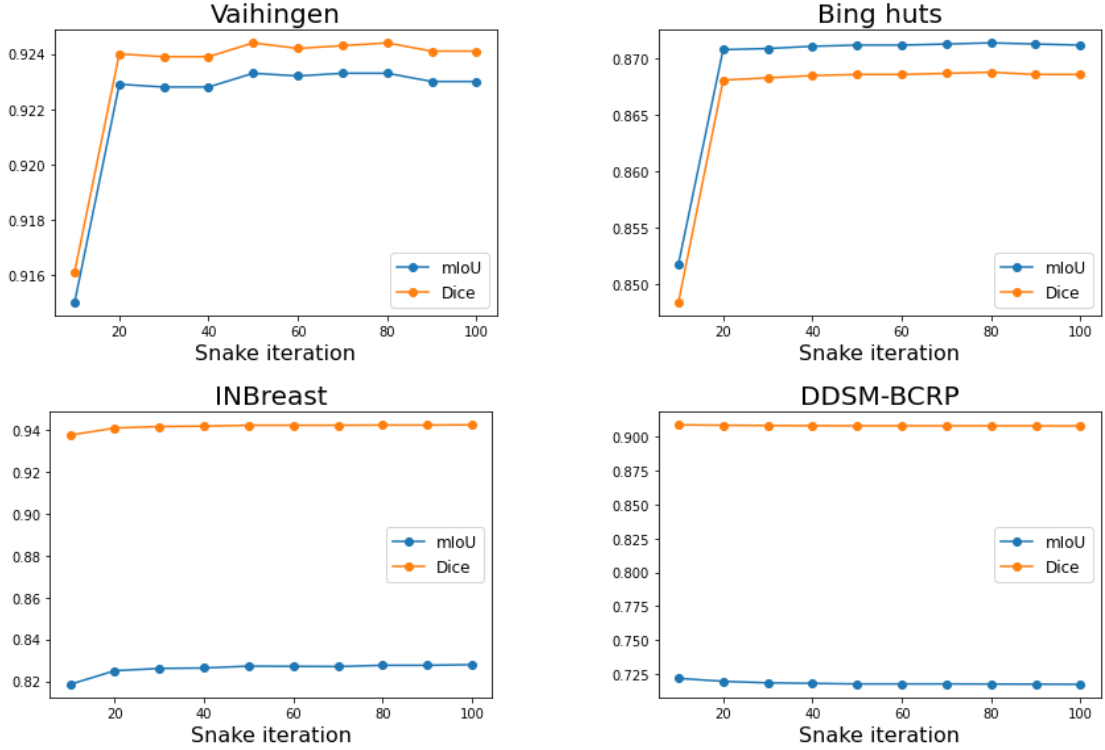

*Figure 10. Evaluation metrics over the number of snake iterations.*

### 5.4.3. Automatic Initialization

Table 6 shows the effect of choosing inscribed or circumscribed circle of the predicted mask as initialization. As can be seen, the proposed method performance slightly depends on the type of automatic initialization circle. However, we choose circumscribed circle for building datasets and inscribed circle for medical datasets as automatic initialization.

*Table 6. The effect of initialization method.*

| Datasets | Inscribed circle | | | Circumscribed circle | | |
|---|---|---|---|---|---|---|
| | mIoU | Dice | BoundF | mIoU | Dice | BoundF |
| Vaihingen | 92.02 | 92.10 | 85.88 | **92.33** | **92.44** | **86.57** |
| Bing huts | 86.81 | 86.47 | 66.27 | **87.12** | **86.86** | **66.91** |
| INBreast | **82.72** | **94.23** | **60.67** | 82.33 | 94.09 | 59.83 |
| DDSM-BCRP | **72.16** | **90.89** | **43.20** | 71.15 | 90.72 | 42.39 |

### 5.4.4. Capture Range

DSAC framework is susceptible to initialization, which is also addressed due to using DVF as external energy. The proposed method provides a wide range of optional initializations. Figure 11 showing mIoU over initialization circle radius indicates the importance of active contour internal energy and kappa term in models capture range. It also suggests the proposed method has the least sensitivity to contour initialization. It is interesting to note that although using LCDVF gives comparable result to our model, it has the most sensitivity to contour initialization.

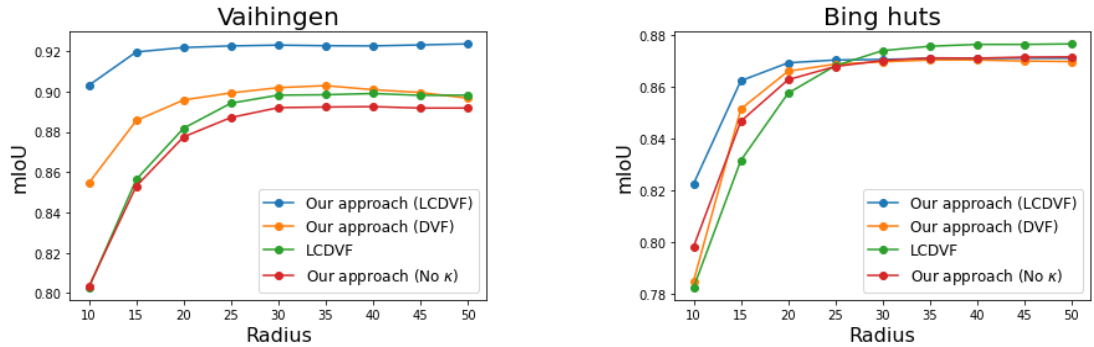

*Figure 11. Model's capture range sensitivity to energy terms.*

Figure 12 demonstrates building segmentation using our approach with three different initializations compared to DSAC. As can be seen, DSAC is highly sensitive to initialization. If the initialization is chosen to be outside the boundary, contour does not converge due to local minima and gradient invisibility. Therefore, the initialization needs to be inside the target and close enough to be able to capture the edges, otherwise contour might converge to local minima inside the target. On the other hand, our method performs much better and converges to edges in three cases with almost the same quality.

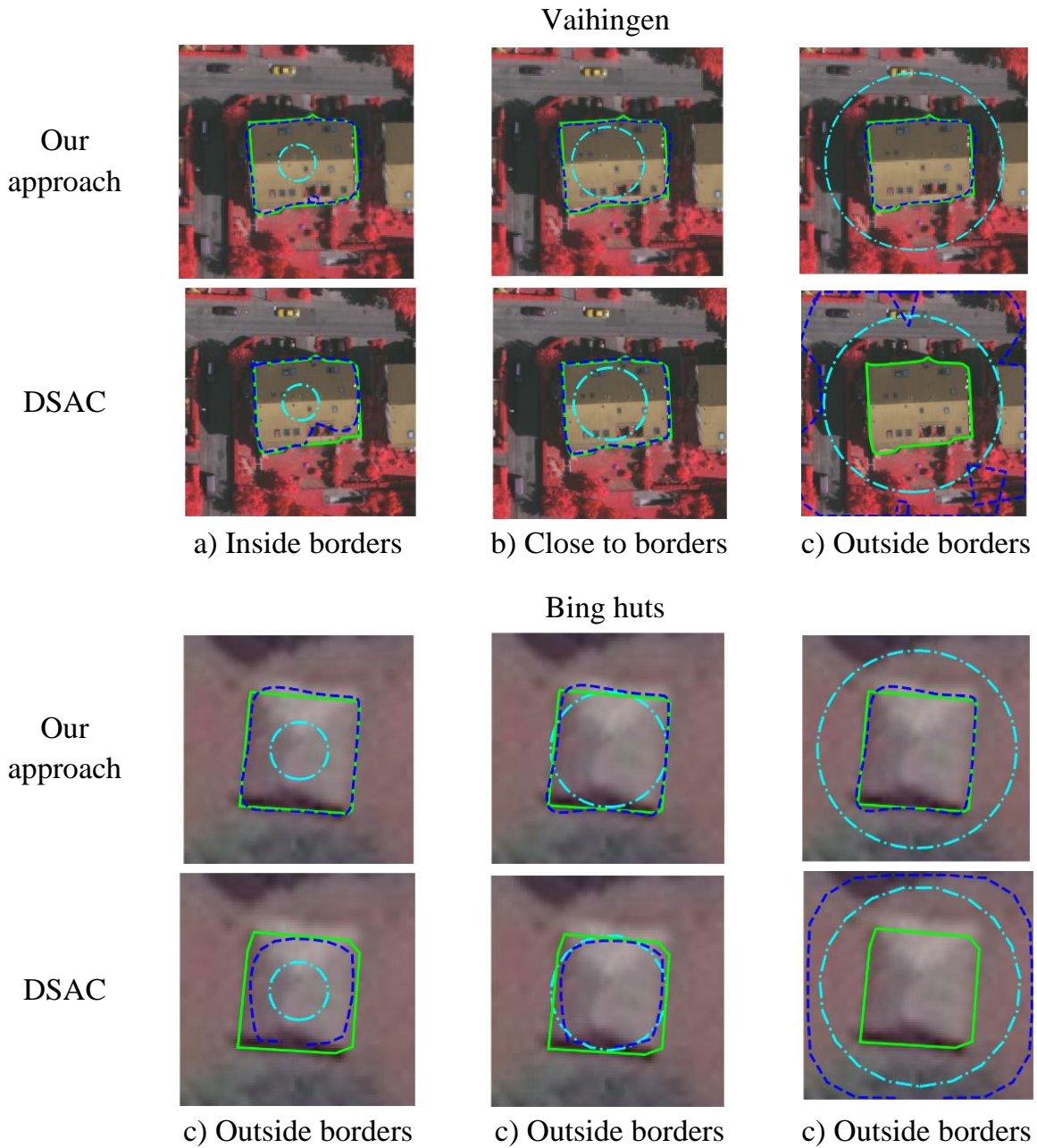

*Figure 12. initialization sensivity. Ground truth, initial contour and the result are in solid green line, dash-dot cyan and dashed blue respectively.*

## 6. Conclusions

In this work we have proposed locally controlled distance vector field in which a vector pointing toward the nearest point on the edge with the magnitude of corresponding distance is learned for each pixel. Active contour parameters and initialization are predicted by a CNN. The method is fully automatic and needs no human supervision. We have tested our approach with four datasets which indicates high capture range and concavity extraction capability in comparison to competing state-of-the-art methods.

# REFERENCES

[1] Kass, M., Witkin, A. and Terzopoulos, D., 1988. Snakes: Active contour models. *International journal of computer vision*, *1*(4), pp.321-331.

[2] Cohen, L.D. and Cohen, I., 1993. Finite-element methods for active contour models and balloons for 2-D and 3-D images. *IEEE Transactions on Pattern Analysis and machine intelligence*, *15*(11), pp.1131-1147.

[3] Xu, C. and Prince, J.L., 1998. Snakes, shapes, and gradient vector flow. *IEEE Transactions on image processing*, *7*(3), pp.359-369.

[4] Sum, K.W. and Cheung, P.Y., 2007. Boundary vector field for parametric active contours. *Pattern Recognition*, *40*(6), pp.1635-1645.

[5] Xie, X. and Mirmehdi, M., 2008. MAC: Magnetostatic active contour model. *IEEE Transactions on pattern analysis and machine intelligence*, *30*(4), pp.632-646.

[6] Wang, T., Cheng, I. and Basu, A., 2009. Fluid vector flow and applications in brain tumor segmentation. *IEEE Transactions on Biomedical Engineering*, *56*(3), pp.781-789.

[7] Cohen, L.D., 1991. On active contour models and balloons. *CVGIP: Image understanding*, *53*(2), pp.211-218.

[8] Marcos, D., Tuia, D., Kellenberger, B., Zhang, L., Bai, M., Liao, R. and Urtasun, R., 2018. Learning deep structured active contours end-to-end. In *Proceedings of the IEEE Conference on Computer Vision and Pattern Recognition* (pp. 8877-8885).

[9] Cheng, D., Liao, R., Fidler, S. and Urtasun, R., 2019. Darnet: Deep active ray network for building segmentation. In *Proceedings of the IEEE Conference on Computer Vision and Pattern Recognition* (pp. 7431-7439).

[10] Gur, S., Shaharabany, T. and Wolf, L., 2019. End to End Trainable Active Contours via Differentiable Rendering. *arXiv preprint arXiv:1912.00367*.

[11] Rupprecht, C., Huaroc, E., Baust, M. and Navab, N., 2016. Deep active contours. *arXiv preprint arXiv:1607.05074*.

[12] Hoogi, A., Subramaniam, A., Veerapaneni, R. and Rubin, D.L., 2016. Adaptive estimation of active contour parameters using convolutional neural networks and texture analysis. *IEEE transactions on medical imaging*, *36*(3), pp.781-791.

[13] Le, T.H.N., Quach, K.G., Luu, K., Duong, C.N. and Savvides, M., 2018. Reformulating level sets as deep recurrent neural network approach to semantic segmentation. *IEEE Transactions on Image Processing*, *27*(5), pp.2393-2407.

[14] Hatamizadeh, A., Sengupta, D. and Terzopoulos, D., 2019. End-to-end deep convolutional active contours for image segmentation. *arXiv preprint arXiv:1909.13359*.

[15] Wang, Z., Acuna, D., Ling, H., Kar, A. and Fidler, S., 2019. Object instance annotation with deep extreme level set evolution. In *Proceedings of the IEEE Conference on Computer Vision and Pattern Recognition* (pp. 7500-7508).

[16] Hatamizadeh, A., Sengupta, D. and Terzopoulos, D., 2020, August. End-to-End Trainable Deep Active Contour Models for Automated Image Segmentation: Delineating Buildings in Aerial Imagery. In *European Conference on Computer Vision* (pp. 730-746). Springer, Cham.

[17] International society for photogrammetry and remote sensing, 2d semantic labeling contest. http://www2.isprs.org/commissions/comm3/wg4/semantic-labeling.html.

[18] Moreira, I.C., Amaral, I., Domingues, I., Cardoso, A., Cardoso, M.J. and Cardoso, J.S., 2012. Inbreast: toward a full-field digital mammographic database. *Academic radiology*, *19*(2), pp.236-248.

[19] Heath, M., Bowyer, K., Kopans, D., Kegelmeyer, P., Moore, R., Chang, K. and Munishkumaran, S., 1998. Current status of the digital database for screening mammography. In *Digital mammography* (pp. 457-460). Springer, Dordrecht.

[20] Ball, J.E. and Bruce, L.M., 2007, August. Digital mammographic computer aided diagnosis (cad) using adaptive level set segmentation. In *2007 29th Annual International Conference of the IEEE Engineering in Medicine and Biology Society* (pp. 4973-4978). IEEE.

[21] Zhu, W., Xiang, X., Tran, T.D., Hager, G.D. and Xie, X., 2018, April. Adversarial deep structured nets for mass segmentation from mammograms. In *2018 IEEE 15th International Symposium on Biomedical Imaging (ISBI 2018)* (pp. 847-850). IEEE.

[22] Li, H., Chen, D., Nailon, W.H., Davies, M.E. and Laurenson, D., 2018. Improved breast mass segmentation in mammograms with conditional residual u-net. In *Image Analysis for Moving Organ, Breast, and Thoracic Images* (pp. 81-89). Springer, Cham.

[23] Singh, V.K., Rashwan, H.A., Romani, S., Akram, F., Pandey, N., Sarker, M.M.K., Saleh, A., Arenas, M., Arquez, M., Puig, D. and Torrents-Barrena, J., 2020. Breast tumor segmentation and shape classification in mammograms using generative adversarial and convolutional neural network. *Expert Systems with Applications*, *139*, p.112855.